\documentclass[10pt,twocolumn,letterpaper]{article}

\usepackage{iccv}              

\definecolor{iccvblue}{rgb}{0.21,0.49,0.74}
\usepackage[pagebackref,breaklinks,colorlinks,allcolors=iccvblue]{hyperref}
\usepackage{graphicx}
\usepackage{amsmath}
\usepackage{amsthm}
\usepackage{booktabs}
\usepackage{algorithm}
\usepackage{algorithmic}
\usepackage{float}

\def\paperID{14203} 
\def\confName{ICCV}
\def\confYear{2025}

\title{InterGSEdit: Interactive 3D Gaussian Splatting Editing with\\ 3D Geometry-Consistent Attention Prior}

\author{Minghao Wen$^{1\;*}$ \quad
Shengjie Wu$^{1}$ \thanks{These authors contributed equally to this work.}\quad
Kangkan Wang$^{2\;\dagger}$  \quad
Dong Liang$^{1} \thanks{Corresponding author.}$ \\
$^{1}$MIIT Key Laboratory of Pattern Analysis and Machine Intelligence, \\
College of Computer Science and Technology, Nanjing University of Aeronautics and Astronautics \\
$^{2}$The Key Lab of Intelligent Perception and Systems for\\ High-Dimensional Information of Ministry of Education, \\
School of Computer Science and Engineering, Nanjing University of Science and Technology \\
{\tt\small \{lanche, wushengjie, liangdong\}@nuaa.edu.cn, wangkangkan@njust.edu.cn}
}
\begin{document}

\maketitle

\begin{abstract}
3D Gaussian Splatting based 3D
editing
has demonstrated impressive performance in recent years.
However, the multi-view editing
 often exhibits significant local inconsistency, especially in areas of non-rigid deformation,
which lead to local artifacts, texture blurring, or semantic 
variations in edited 3D scenes.
We also found that the existing editing methods, which rely entirely on text prompts make the editing process a "one-shot deal", making it difficult for users to control the editing degree flexibly.
In response to these challenges, we present InterGSEdit, a novel framework for high-quality 3DGS 
editing via interactively selecting key views with users' preferences.
We propose a CLIP-based Semantic Consistency Selection (CSCS) strategy to adaptively screen a group of semantically consistent reference views for each user-selected key view.
Then, the cross-attention maps derived from the reference views are used in a weighted Gaussian Splatting 
unprojection
to construct the 3D Geometry-Consistent Attention Prior ($GAP^{3D}$).
We project $GAP^{3D}$ to obtain 3D-constrained attention, which are fused with 2D cross-attention via 
Attention Fusion Network (AFN). AFN employs an adaptive attention strategy that prioritizes 3D-constrained attention for geometric consistency during early inference, and gradually prioritizes 2D cross-attention maps in diffusion for fine-grained 
features during the later inference.
Extensive experiments demonstrate that InterGSEdit achieves state-of-the-art performance, delivering consistent, high-fidelity 3DGS editing with improved user experience.
\end{abstract} 
\section{Introduction}
\label{sec:Introduction}

\begin{figure*}[t]
    \hsize=\textwidth 
    \centering
    \includegraphics[width=.92\textwidth]{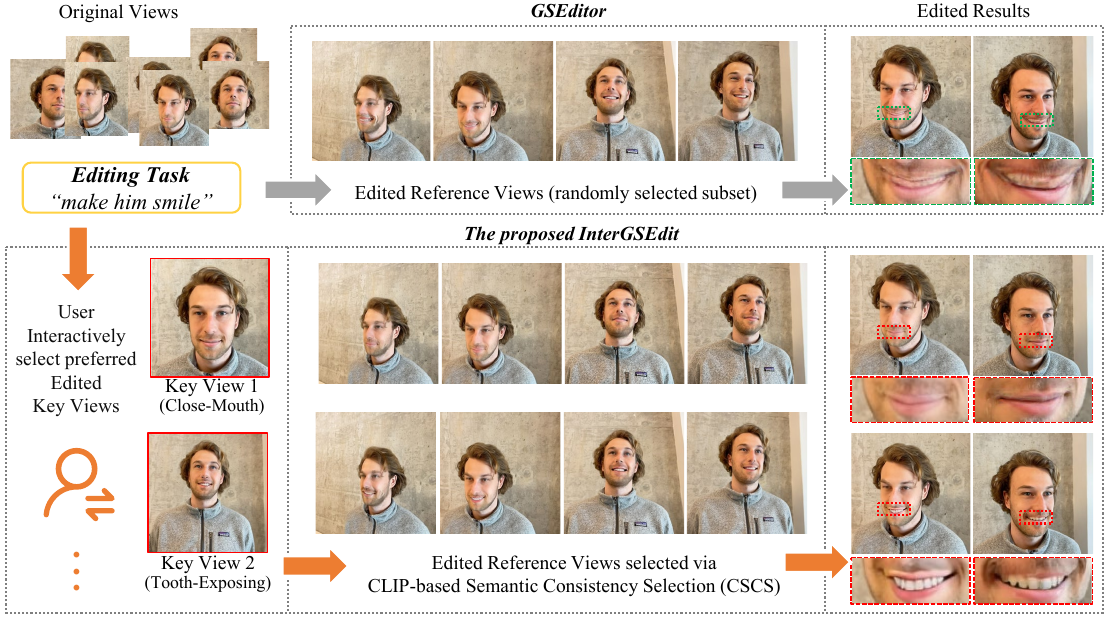}
    \vspace{-.3cm}
    \caption{\textbf{Results of a ``make him smile'' editing task.}     GSEditor~\cite{chen2023gaussianeditorswiftcontrollable3d} randomly select a subset of edited views as reference view to generate the 3DGS results, leading to blur synthesis due to multi-view inconsistency, such as tooth artifacts in this example.
    In contrast, our InterGSEdit framework allows the user to select the preferred key views from the edited views to guide the multi-view editing.
    Here, we illustrate two key views with distinct smiling characteristics (close-mouth smile and tooth-exposing smile). 
    The edited results exhibit geometric consistency with the key views, more natural textures, and finer local details compared to GSEditor.
    }
    \vspace{-.3cm}
    \label{fig1:introduction}
\end{figure*}

The development of scene representation models, such as Neural Radiance Fields (NeRF)~\cite{mildenhall2020nerfrepresentingscenesneural} and 3D Gaussian Splatting (3DGS)~\cite{kerbl20233dgaussiansplattingrealtime}, has made 
scene reconstruction and rendering both efficient and practical. These advances have significantly boosted the research of 3D scene editing, and text-guided diffusion editing methods
~\cite{chen2023gaussianeditorswiftcontrollable3d,wang2024gaussianeditor,tang2024dreamgaussiangenerativegaussiansplatting,wu2024gaussctrlmultiviewconsistenttextdriven,wang2024viewconsistent3deditinggaussian,chen2024dgedirectgaussian3d,rombach2022highresolutionimagesynthesislatent,brooks2023instructpix2pixlearningfollowimage} attracts increasing attention in recent years. Text-guided diffusion methods typically first edit multi-view images rendered from a 3D scene model (i.e., 3DGS) according to the target text description, and subsequently refine the 3D scene model with the modified images to generate 3D editing results. Notably, 3DGS-based scene editing~\cite{wang2024gaussianeditor,chen2023gaussianeditorswiftcontrollable3d,chenproedit,wang2024viewconsistent3deditinggaussian,wu2024gaussctrlmultiviewconsistenttextdriven,chen2024dgedirectgaussian3d} has achieved remarkable success in rigid editing tasks such as style transfer and appearance adjustment. However, it is still challenging to edit non-rigid scenes due to more complicated and arbitrarily varying deformations. 

The current mainstream approaches~\cite{chen2024dgedirectgaussian3d,wang2024viewconsistent3deditinggaussian,wu2024gaussctrlmultiviewconsistenttextdriven,luo2023latent} rely on different 3D constraint techniques, such as epipolar constraints~\cite{chen2024dgedirectgaussian3d}, consistent 3DGS fine-tuning~\cite{wang2024viewconsistent3deditinggaussian},  or depth map guidance~\cite{wu2024gaussctrlmultiviewconsistenttextdriven}, to enforce cross-view consistency.
Although these strategies work well for rigid editing tasks like style transfer and appearance modification, there are obvious limitations in non-rigid scene editing like faces with changing expressions.
As illustrated in \cref{fig1:introduction}, in a ``smile'' editing task, the edited faces in different views may exhibit varying appearances due to non-rigid deformations in the aspects of smile intensity and tooth visibility. Refining 3DGS face with these inconsistent face images will result in blurred artifacts or distorted geometry. 
The primary reason for this problem is that the textual guidance in diffusion-based editing is inherently ambiguous and cannot be specified in detail to achieve consistent and precise editing. The non-rigid editing easily leads to inconsistent features across different views during the diffusion process, making it difficult to ensure 3D geometric consistency and high-quality synthesis.

We also found that the existing 3DGS editing methods that are entirely instructed by text prompts suffer from ``one-shot deal" problem in the editing process, making it difficult for users to flexibly control the editing degree. The fine-grained representation capability of the text prompts is limited by the diffusion network designation, training strategies, training sample quality, as well as users' experience, and so on. Moverover, it is hard for the text prompts to specify fine-grained semantics due to the inherently ambiguous nature, and 
this linguistic uncertainty leads to variations in editing features across views, resulting in inconsistency and artifacts. In practice, achieving the 3D editing that users desire remains challenging due to the concise text prompts and the constraints of pre-trained diffusion networks.

To alleviate the above problems in 3D editing tasks,
we propose a user interactive method, InterGSEdit, which builds 3D Geometry-Consistent Attention Prior ($GAP^{3D}$) based on a user-selected key view and ensures that the final edited results are closely consistent with the key view in both appearance and geometry. Specifically, we first perform initial editing on the rendered views, producing edited results that may exhibit discrepancies. Then, based on a key view chosen by the user, 
our CLIP-based
Semantic Consistency Selection (CSCS) treats the key view as an anchor to select reference views with their corresponding similarity weights that are computed from the embeddings of edited images. Finally, we construct the 3D Geometry-Consistent Attention prior ($GAP^{3D}$) on the 3DGS by performing a weighted unprojection of the cross-attention maps in the diffusion~\cite{chen2023gaussianeditorswiftcontrollable3d} from these selected reference views.

Then, we incorporate the constructed 3D attention prior into the same diffusion model to achieve consistent editing among different views. We project $GAP^{3D}$ to each edited image to form a 3D-constrained attention map. To incorporate 3D geometric constraints during the denoising process, directly replacing the 2D cross-attention with the 3D-constrained attention can result in the decay of the editing details.
To overcome this, we propose an adaptive cross-dimensional Attention Fusion Network (AFN) that employs a learnable gating module to adaptively fuse 2D cross-attention maps with 3D attention features. 
AFN ensures that during the early stages of editing, 3D-constrained attention is preferred to maintain 3D geometric consistency, while in the later stages, 2D cross attention is focused more on to preserve the fidelity of editing details, thereby achieving both structural consistency and detail recovery.

Based on InterGSEdit with these essential modules, we achieve high-quality 3D scene editing in both non-rigid tasks (e.g., facial editing), and rigid tasks (e.g., appearance modification and style transfer). 
Users can choose a key view to guide the editing process and obtain the final 3D editing results that satisfy their demands on editing. Compared to existing methods, our approach is more flexible and effective to obtain view-consistent editing results and achieves 
high-quality synthesis and geometry.
Our main contributions can be summarized in three aspects:
\begin{itemize}
    \item We propose a user-interactive way of improving the user experience for fine-grained 3D editing and constructing 3D Geometry-Consistent Attention Prior ($GAP^{3D}$). After interactively selecting the key views, we employ our CLIP-based Semantic Consistency Selection (CSCS) strategy to obtain qualified reference views and then unproject their cross-attention maps 
    to build $GAP^{3D}$.
    \item We propose an adaptive cross-dimensional Attention Fusion Network (AFN) to adaptively and dynamically fuse the $GAP^{3D}$ with the 2D cross-attention map during the inference stages, supporting both multi-view consistency and fine-grained detail recovery.
    \item We propose a 3DGS editing framework, InterGSEdit, which leverages CSCS and AFN to corporately generate reliable $GAP^{3D}$ and then achieve multi-view consistent and high-quality editing in the diffusion. Our approach demonstrates state-of-the-art performance across various scenarios on public datasets~\cite{haque2023instructnerf2nerfediting3dscenes}.
    
\end{itemize}
\section{Related works}
\label{sec:Related works}
\textbf{Image Editing with Diffusion Models.} 
Due to the scarcity of large-scale 3D training datasets for 3D scene editing, current methods mainly utilize 2D image editing and then estimate 3D models from edited images. Among these approaches, denoising diffusion probabilistic models (DDPM)~\cite{ho2020denoisingdiffusionprobabilisticmodels} are the most popular. Stable Diffusion~\cite{rombach2022highresolutionimagesynthesislatent} performs diffusion in a latent space, which reduces the scale of the U-Net architecture and improves the performance of diffusion models.
To control the editing process, ControlNet~\cite{zhang2023addingconditionalcontroltexttoimage}  introduces control guidance such as scribbles or depth maps. GLIDE~\cite{nichol2022glidephotorealisticimagegeneration} leverages CLIP~\cite{radford2021learningtransferablevisualmodels} features for image editing, while DreamBooth~\cite{ruiz2023dreamboothfinetuningtexttoimage} focuses on personalized editing. Additionally, DragDiffusion~\cite{shi2024dragdiffusionharnessingdiffusionmodels} and Drag-a-Video~\cite{teng2023dragavideononrigidvideoediting} employ interactive mouse guidance in the editing process. Recent methods mainly focus on text-driven guidance, which is very relevant to our work. For instance, InstructPix2Pix~\cite{brooks2023instructpix2pixlearningfollowimage} incorporates textual conditions in Stable Diffusion, and achieves high-quality text-driven editing by training on large-scale synthetic datasets. InfEdit~\cite{xu2023inversionfreeimageeditingnatural} enables faithful editing for both rigid and non-rigid semantic changes using a denoising diffusion consistent model.
\\
\textbf{3D Editing.}
Early 3D editing methods~\cite{michel2021text2meshtextdrivenneuralstylization,chen2022tangotextdrivenphotorealisticrobust,siddiqui2024meshgpt,zhang2024towards} are primarily based on traditional 3D models by synthesizing or refining meshes using text guidance. These mesh-based methods often focus on modeling a specific object rather than general scenes, resulting in a limited scope for editing. Furthermore, the editing is typically confined to modifications or optimizations on color and texture, which constrains the applicability of such editing. 
With the advent of Neural Radiance Fields (NeRF)~\cite{mildenhall2020nerfrepresentingscenesneural}, subsequent approaches~\cite{wang2022clipnerftextandimagedrivenmanipulation,Gordon_2023,yuan2022nerf,he2024customize} edit implicit scene representations, such as methods~\cite{hong2022avatarclip,cuilam3d,wu2025neural,wu2024unique3d} tailored for 3D character models. These methods have expanded the editing target to a relatively large scene. However, constrained by the editing capability of the guidance prior, the modifications remain confined to aspects such as color, texture, or rotation/scaling of objects. To alleviate the high computation in NeRF optimization, DreamEditor~\cite{zhuang2023dreameditortextdriven3dscene} utilizes view-specific masks to constrain the editing area. Also, Score Distillation Sampling ~\cite{alldieck2024score} is introduced in~\cite{kamata2023instruct3dto3dtextinstruction,sella2023voxetextguidedvoxelediting,cheng2024progressive3dprogressivelylocalediting} to enhance training speed and editing accuracy.
More recently, diffusion-based methods edit 3D radiance fields by editing images rendered from multiple viewpoints and optimizing the underlying 3D model. With the development of the diffusion model, NeRF-based editing methods~\cite{haque2023instructnerf2nerfediting3dscenes} incorporate diffusion models to achieve high-quality editing. The powerful ability of diffusion models enables complex 3D editing operations such as style transfer, or changing scene features such as character movements or attributes.
\\
\textbf{Multi-View Consistent 3D Gaussian Editing.} 
Recent studies~\cite{zhou2024feature3dgssupercharging3d,chen2023gaussianeditorswiftcontrollable3d,wang2024gaussianeditor} improve accuracy, efficiency, and controllability by integrating 3DGS into 3D scene editing. 
Based on 3DGS, several methods are proposed to address multi-view inconsistency problem in 3D scene editing, especially for rigid editing tasks. Current methods~\cite{chen2024dgedirectgaussian3d,wang2024viewconsistent3deditinggaussian,wu2024gaussctrlmultiviewconsistenttextdriven} focus on incorporating 3D geometric constraints across multiple views to improve consistency by manipulating attention maps during the 2D editing stage. DGE~\cite{chen2024dgedirectgaussian3d} uses epipolar constraints to build pixel correspondences and propagate point features of the key view to the general view. VcEdit~\cite{wang2024viewconsistent3deditinggaussian} reversely projects the attention maps to 3D Gaussians and then renders refined attention maps to achieve multi-view attention unification. GaussCtrl~\cite{wu2024gaussctrlmultiviewconsistenttextdriven} minimizes the view differences by editing all views in parallel. ProEdit~\cite{chenproedit} decomposes an editing task into multiple subtasks to control the feasible output space, thereby reducing inconsistencies in the final results. These methods~\cite{chen2024dgedirectgaussian3d,wang2024viewconsistent3deditinggaussian} improve the consistency of multiple views to some degree, but they can lead to artifacts in non-rigid situations when the differences in 2D editing results among views are obvious. We allow users to select key views to ensure that the edited results meet their expectations, and also utilize an adaptive cross-dimensional attention fusion to maintain both view consistency and detail recovery.


\begin{figure*}[!]
    \centering
    \includegraphics[width=1\textwidth]{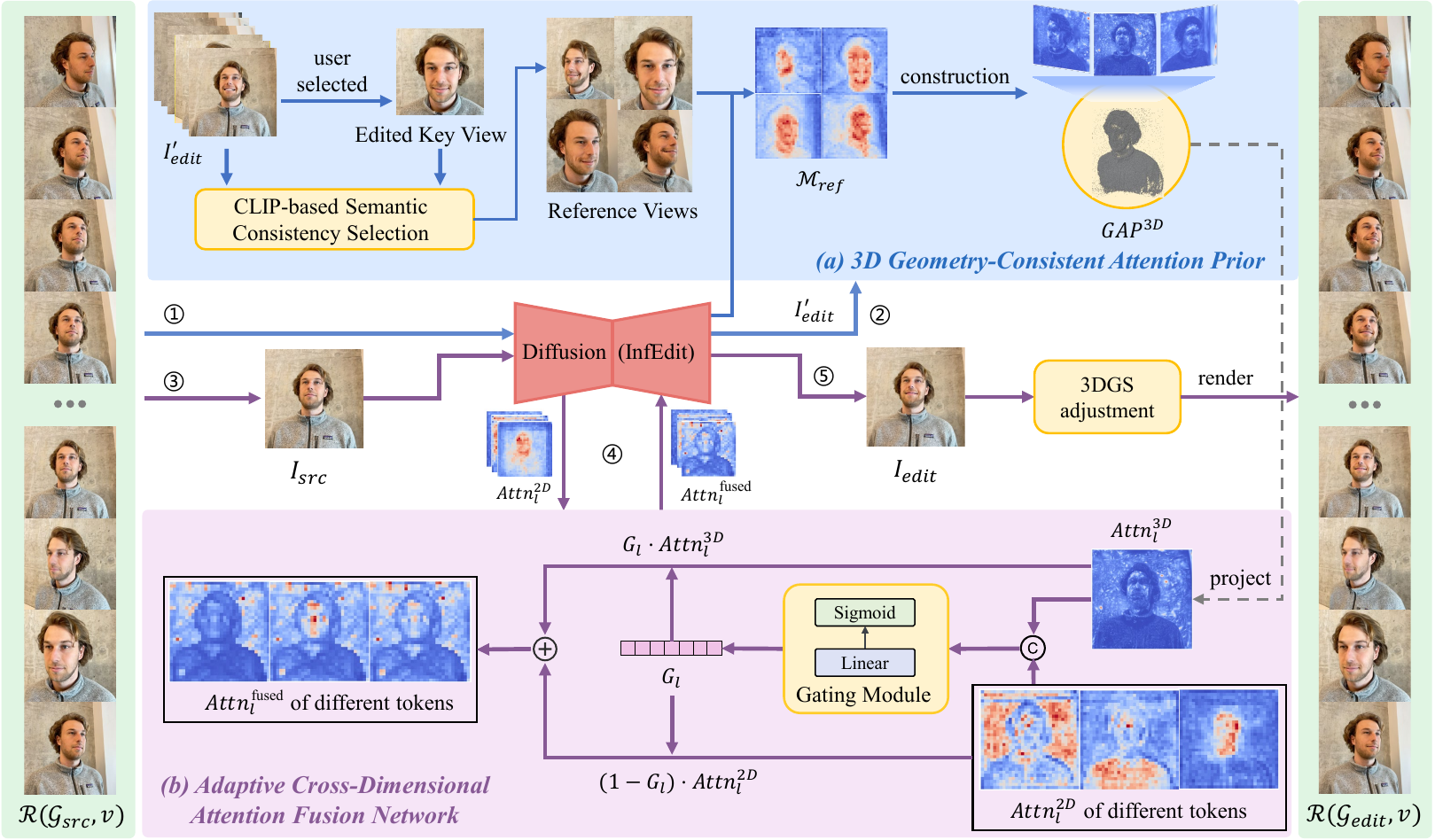} %
    \caption{\textbf{Our InterGSEdit framework} 
    mainly comprises two components. (a) 3D Geometry-Consistent Attention Prior ($GAP^{3D}$) Construction.  User-specified key views serve to select reference views with high semantic consistency. We employ a CLIP-based Semantic Consistency Selection (CSCS) strategy to select semantically consistent reference views and then utilize their cross-attention maps to construct $GAP^{3D}$ (\cref{subsec:first-stage}).
    (b) Adaptive Cross-Dimensional Attention Fusion Network (AFN). We introduce a dynamic gating module to fuse 2D cross-attention with $GAP^{3D}$ based on gating factor $G_l$,  (\cref{subsec:second-stage}), and then feed the fused results into diffusion editing.}
    \label{fig2:framework}
    \vspace{-.3cm}
\end{figure*}

\section{Preliminary}
\label{sec:Preliminary}
\textbf{3D Gaussian Splatting}~\cite{kerbl20233dgaussiansplattingrealtime}. In 3DGS, each Gaussian, represented as \(G\), is characterized by its mean \(\mu \in \mathbb{R}_{}^{3}\), covariance matrix $C$, associated color \(c \in \mathbb{R}_{}^{3}\), and opacity \(\alpha \in \mathbb{R}\). The covariance matrix $C$ can be decomposed into a scaling matrix \(S \in \mathbb{R}_{}^{3}\) and a rotation matrix \(R \in \mathbb{R}_{}^{3 \times 3}\), with \(C = RSS_{}^{T}R_{}^{T}\). A Gaussian centered at \(\mu\) is expressed as \(G(x) = exp(-\frac{1}{2}x_{}^{T}C_{}^{-1}x)\), where \(x\) denotes the displacement from \(\mu\) to a point in 3D space. In the splatting rendering, the pixel color \(c\) is rendered by blending all sampled 3D points along the ray emitted from this pixel:
\begin{equation}
    c=\sum_{i}c_{i}\alpha_{i}G(x_{i})\prod_{j=1}^{i-1}(1-\alpha_{j}G(x_{j})).
\end{equation}

\noindent \textbf{Diffusion-based 3DGS Editing.}
Some methods~\cite{chen2024dgedirectgaussian3d,wang2024viewconsistent3deditinggaussian,wu2024gaussctrlmultiviewconsistenttextdriven,chen2023gaussianeditorswiftcontrollable3d} render the 3D scene model $\mathcal{G}_{\text{src}}$ from multiple viewpoints $v$ to obtain 2D source images $I_{\mathrm{src}}^v$. A diffusion-based editor then modifies these 2D images into edited images $I_{\mathrm{edit}}^v$ according to the prompt $y$. By comparing the rendered images with the edited images by diffusion, an editing loss $\mathcal{L}_{\mathrm{Edit}}$ can be defined that measures how closely the rendered outputs of the 3D model align with the edited results guided by diffusion.
When optimizing across all rendered viewpoints $v$, the objective function is defined as:
\begin{equation}\label{editloss}
    \mathcal{G}^{\text{edit}} = \arg\min_{\mathcal{G}} \sum_{v \in \mathcal{V}} \mathcal{L}_{\text{Edit}}(\mathcal{R}(\mathcal{G},v),I_{\mathrm{edit}}^v),
\end{equation}
where $\mathcal{R}(\mathcal{G}, v)$ denotes  differential rendering the edited $\mathcal{G}$ from the camera viewpoint $v$.

\section{Methodology}
Our proposed InterGSEdit aims to interactively edit a 3D scene with 3D Geometry-Consistent Attention Prior ($GAP^{3D}$) guided diffusion model. An illustration of our framework is shown in \cref{fig2:framework}.
First, the original 3D model $\mathcal{G}_{src}$ is rendered from multiple views $v$ to obtain source images $I_{src}=\mathcal{R}(\mathcal{G},v)$, and $I_{src}$ are initially edited to obtain $I'_{edit}$ with diffusion.
Then, as described in \cref{subsec:first-stage}, we introduce a CLIP-based Semantic Consistency Selection (CSCS), which leverages a user-selected key view as a semantic anchor to obtain semantically consistent reference views. 
These reference views, along with their associated weights, are used to construct $GAP^{3D}$.
Subsequently, as described in \cref{subsec:second-stage}, we perform 3D-constrained editing on $I_{src}$ using a cross-dimensional Attention Fusion Network (AFN).
This network employs a 3D-2D attention fusion module to inject the 3D-constrained attention, derived from $GAP^{3D}$ projection, into the denoising process of the diffusion model. Through this diffusion editing, we obtain multi-view consistent images and the edited 3DGS.

\subsection{3D Geometry-Consistent Attention prior}
\label{subsec:first-stage}
\subsubsection{CLIP-based Semantic Consistency Selection}
The text prompts are inherently ambiguous to specify fine-grained semantics, and 
this linguistic uncertainty leads to variations in editing features across different views, resulting in 3D geometric inconsistency, blur outputs, and artifacts. 
To address this issue, we anchor the ambiguous semantics by allowing users to select an edited image as a key view, and edit the 3D scene according to the content of the key view. We propose a CLIP-based Semantic Consistency Selection (CSCS) strategy for dynamically selecting semantically consistent reference views, 
which quantifies the editing similarity by measuring distances in a cross-modal embedding space.
Specifically, we first use the CLIP image encoder $E_{\mathrm{CLIP}}^{\mathrm{img}}(\cdot)$~\cite{radford2021learningtransferablevisualmodels} to obtain image embeddings of the edited key view $I_{\text{edit}}^{\text{key}}$ and its corresponding original image $I_{\text{src}}^\mathrm{key}$, and compute their difference in editing content:

\begin{equation}
    \Delta I_{\text{key}} = E_{\mathrm{CLIP}}^{\mathrm{img}}(I_{\mathrm{edit}}^{\mathrm{key}}) - E_{\mathrm{CLIP}}^{\mathrm{img}}(I_{\mathrm{src}}^\mathrm{key}),
\end{equation}
which captures the changes in the editing direction relative to the original image. Similarly, we use the CLIP text encoder $E_{\mathrm{CLIP}}^{\mathrm{txt}}(\cdot)$ to obtain text embeddings of the edit prompt $T_{edit}$ and original text $T_{src}$, and compute their difference as:
\begin{equation}
    \Delta T = E_{\mathrm{CLIP}}^{\mathrm{txt}}(T_{\mathrm{edit}}) - E_{\mathrm{CLIP}}^{\mathrm{txt}}(T_{\mathrm{src}}),
\end{equation}
which reflects the changes in the textual description corresponding to the editing operation. 
Since $\Delta T$ remains constant in denoising, it serves as a fixed reference vector representing the purpose of textual editing. We then compute the cosine similarity between $\Delta I_{\text{key}}$ and $\Delta T$ to obtain the alignment score for the key view:
\begin{equation}
    s_{\text{key}} = D(\Delta I_{\text{key}}, \Delta T),
\end{equation}
where $D$ denotes the cosine distance. 
\begin{figure}[H]
    \centering
    \includegraphics[width=0.5\textwidth]{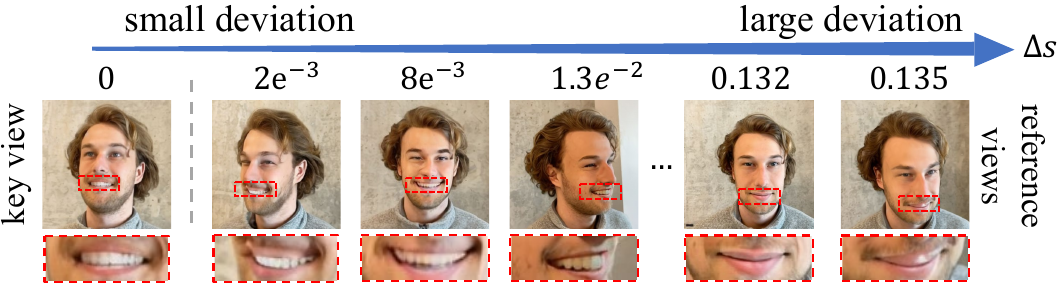}
     \vspace{-.3cm}
    \caption{\textbf{Visualization of CLIP-based Semantic Consistency Selection.} The left image is the key view, whose editing direction is quantified by its alignment score $s_{\text{key}}$. Reference views with lower deviation value $\Delta s_v=|s_v - s_{\text{key}}|$ indicate that their editing changes are more consistent with the key view.}
    \label{fig:example_image}
    \vspace{-.3cm}
\end{figure}
For a view $v$, we compute its alignment score as $s_v=D(\Delta I_v, \Delta T)$.
Using $s_{\mathrm{key}}$ as an anchor, we can calculate deviation value $\Delta s_v=|s_v - s_{\text{key}}|$. The $s_v$ exhibits editing changes that are consistent with the key view under the same textual guidance if $\Delta s_v$ is close to $0$. 
We select the top-$K$ views with minimal deviation:
\begin{equation}
    V_{\mathrm{ref}} = \{ v_{1}, v_{2}, \ldots, v_{K} \},
\end{equation}
where $\Delta s_{v_{1}} \leq \Delta s_{v_{2}} \leq \cdots \leq \Delta s_{v_{K}}$.
However, applying a hard threshold to select the top-$K$ reference views may exclude true similar views. These views closely resemble the key view in editing features but fall just outside the top-$K$ ranking. Such exclusions may disrupt the consistency of multi-view editing. 
To improve selection robustness, we introduce an adaptive weight assignment. It gives higher weights to views with greater similarity while still retaining those with slightly lower similarity if they provide useful reference information. It dynamically adjusts view weights using an exponential decay:
\begin{equation}
    w_v = \exp\left(-\gamma \Delta s_v\right),
    \label{equa:weight of view}
\end{equation}
where $w_v$ is the assigned weight for view $v$, $\exp$ denotes the exponential function, and $\gamma$ is a temperature coefficient controlling the influence of alignment deviation. If the alignment score of a candidate view $s_v$ is close to $s_{\text{key}}$, its weight approaches 1, indicating a significant selection. As the alignment score deviates from $s_{\text{key}}$, the weight rapidly decreases, weakening its probability in the selection. This exponential decay strategy prevents useful information loss from hard-threshold filtering and retains those with slightly lower similarity, which leads to a smoother and more robust reference view selection.
\begin{figure}
    \centering
    \includegraphics[width=0.85\linewidth]{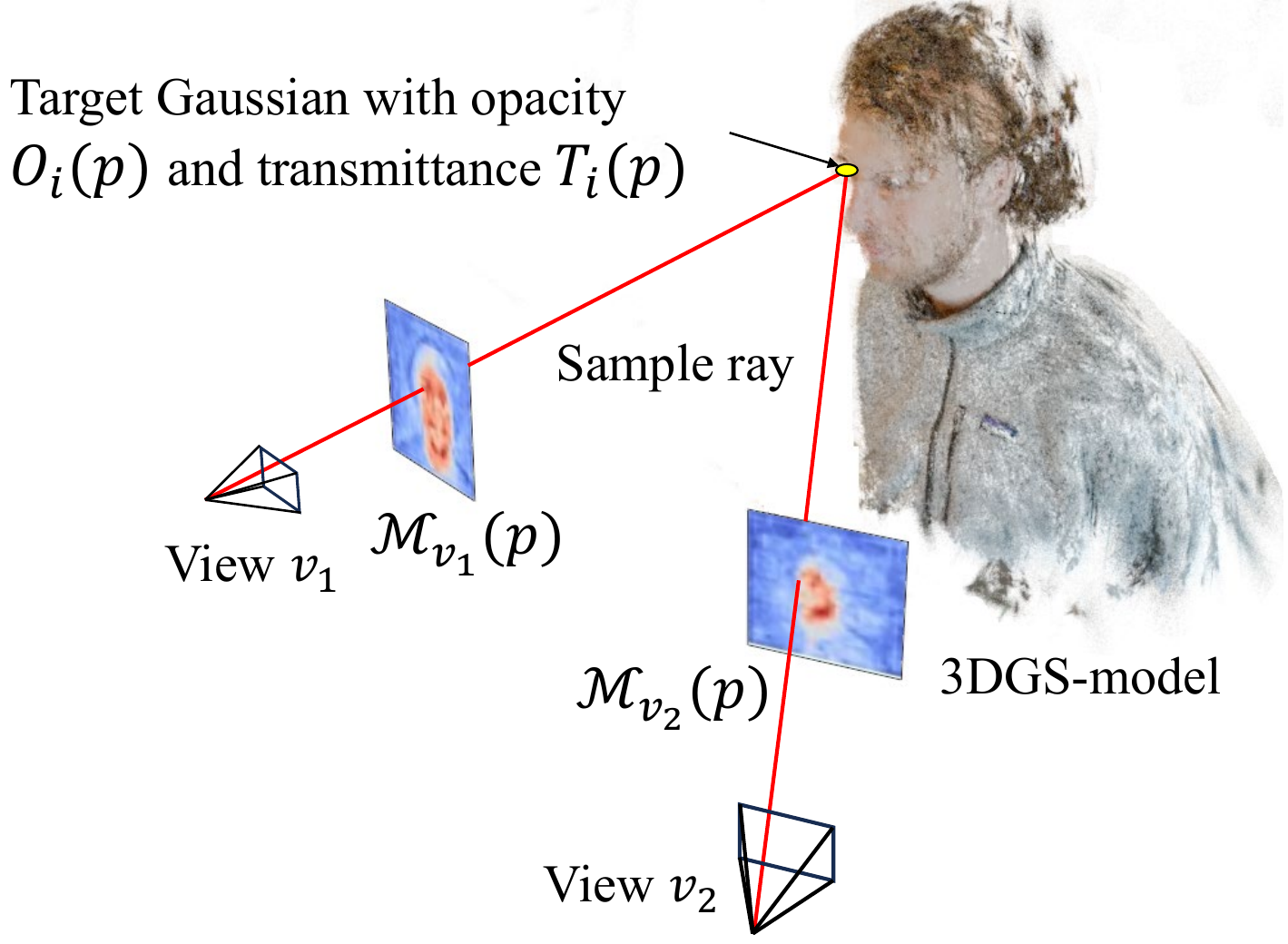}
    \caption{\textbf{Illustration of $GAP^{3D}$ construction.}  Based on the attention map $\mathcal{M}_v$ corresponding to the view $v$ and the associated camera pose, precise correspondences between image pixels and GS points are established. Each view has a similarity weight $w_v$ that is calculated from \cref{equa:weight of view} to construct the 3D attention prior. Here, only two views are shown for illustration.}
    \label{fig:gs-attn}
    \vspace{-.3cm}
\end{figure}
\subsubsection{$GAP^{3D}$ Construction}
In the following, we describe the method for constructing the 3D Geometry-Consistent Attention Prior $GAP^{3D}$.
As illustrated in \cref{fig:gs-attn}, given cross-attention maps $\mathcal{M}_v$ from viewpoint $v$, we first recover the 3DGS scene from reference views and then compute an attention score for each Gaussian with inverse rasterization rendering of 3DGS. Specifically, for each Gaussian point $i$, the attention score $GAP^{3D}(i)$ for Gaussian point $i$ is computed as:
\begin{equation}
    GAP^{3D}(i)=\sum_{v \in V_{ref}}\frac{w_v}{\sum_m w_m} \cdot\sum_{p}\mathcal{M}_v (p) \cdot O_{i}(p) \cdot T_{i}(p),
\end{equation}
where $\mathcal{M}_v (p)$ is the attention score at pixel $p$ in view $v$, $O_{i}(p)$ represents the opacity from Gaussian $i$ to pixel $p$, and $T_{i}(p)$ denotes the transmittance from pixel $p$ to Gaussian $i$~\cite{chen2023gaussianeditorswiftcontrollable3d}.
The normalization factor $\frac{w_v}{\sum_m w_m}$ ensures that, among $m$ reference views related to Gaussian point $i$, the contribution of view $v$ is weighted according to its similarity weight.
This weighting strategy on 2D attention maps establishes a 3D consistency attention prior, enhancing the consistency and quality of subsequent editing.

\subsection{Cross-Dimensional Attention Fusion Network}
\label{subsec:second-stage}
Given a viewpoint $v$, we can obtain a 3D-constrained attention map formulated as $Attn^{\text{3D}}=\mathcal R(GAP^{3D}, v)$ by projecting 3D attention prior to the 2D image space. By injecting 3D-constrained attention into UNet~\cite{ronneberger2015u} of the diffusion model, we can integrate 3D geometry constraints across multiple attention layers.
Here, we design a cross-dimensional Attention Fusion Network (AFN) that fuses 3D-constrained attention $Attn^{\text{3D}}$ with the 2D cross-attention obtained in the editing process. This method can facilitate the stability and consistency of editing features throughout the denoising process. 
Specifically, after the cross-attention map $Attn_l^{\text{2D}}$ is generated at layer $l$ within the diffusion model~\cite{rombach2022highresolutionimagesynthesislatent,brooks2023instructpix2pixlearningfollowimage,luo2023latent}, we project 3D attention prior to a 3D-constrained attention map $Attn_l^{3D}$ and fuse with the cross-attention map $Attn_l^\text{2D}$.

Directly replacing 3D-constrained attention with 2D cross-attention may lead to feature degradation during editing diffusion. To address this problem, AFN adaptively adjusts the influence of 3D-constrained attention at different stages with a dynamic gated fusion mechanism.
Specifically, we introduce a gating factor $G_l$ to fuse the features of $Attn_l^{\text{3D}}$ and $Attn_l^{\text{2D}}$ dynamically.
To ensure that the model emphasizes 3D geometric consistency during the initial inference phase and gradually focuses on editing details later, we design a linearly decaying dynamic bias term given by $\gamma(t)=\alpha (1-\frac{t}{T})$, where $\alpha$ is a constant, $t$ denotes the current iteration, and $T$ is the total number of iterations.
\begin{equation}
    G_l = \sigma \Bigl( W_l \cdot \bigl[ Attn_l^{\text{2D}} ; Attn_l^{\text{3D}} \bigr] + \gamma(t) \Bigr),
\end{equation}
\begin{equation}
    Attn_l^{\text{fused}} = G_l \cdot Attn_l^{\text{3D}} + (1 - G_l) \cdot Attn_l^{\text{2D}},
\end{equation}
where $W_l$ is a learnable weight matrix for the attention fusion at different layers,
and $\sigma$ is the sigmoid function mapping the gating factor $G_l$ into the interval $(0,1)$.
We further incorporate a KL divergence constraint to enhance training stability and precisely regulate the attention fusion process between $Attn^{\text{3D}}$ and $Attn^{\text{2D}}$. The total optimization objective is defined as:
\begin{equation}
    \mathcal{L}_{\text{total}} = \lambda_{\text{2D}} \mathcal{L}_{\text{Edit}} + \lambda_{\text{3D}} \mathcal{L}_{\text{KL}} \Bigl( Attn^{\text{3D}} \parallel Attn^{\text{2D}} \Bigr),
\end{equation}
where $\mathcal{L}_{\text{Edit}}$ represents the editing loss as the diffusion objective in \cref{editloss}, and 
$\lambda_{\text{2D}}$ and $\lambda_{\text{3D}}$ are hyperparameters for two loss terms. $\mathcal{L}_{\text{KL}}$ enforces distributional consistency between the 2D cross-attention $Attn^{\text{2D}}$ and 3D-constrained attention $Attn^{\text{3D}}$.
This constraint forces the $Attn^{\text{2D}}$ distribution to converge to the $Attn^{\text{3D}}$ distribution, thus ensuring that the final generation results are geometry-consistent.

To progressively adjust the attention fusion during inference, we adopt a dynamic regulation strategy that simultaneously optimizes the Attention Fusion Network (AFN) and the scene 3DGS in the denoising process. 
Specifically, the weights $W_{l}$ and a Gating Module are learned to obtain the gating factor $G_l$ for the attention fusion, while a KL divergence constraint is used to regulate the fusion process.
In the early stages, the model is encouraged to prioritize geometric consistency by setting a larger weight $\lambda_{\text{3D}}$ of the KL loss, which forces AFN to adjust $Attn^{\text{2D}}$ to better align with $Attn^{\text{3D}}$.
As denoising process proceeds and the geometric information stabilizes, by gradually reducing $\lambda_{\text{3D}}$, the model is enforced to adjust $G_l$ so that $Attn^{\text{2D}}$ contributes more to the denoising guided by text prompts, ensuring fine editing details are restored.
Consequently, our approach achieves the dual objectives of enforcing structural consistency during early training and progressively emphasizing detail recovery.
\begin{figure*}[t!]
    \centering
    \includegraphics[width=1\textwidth]{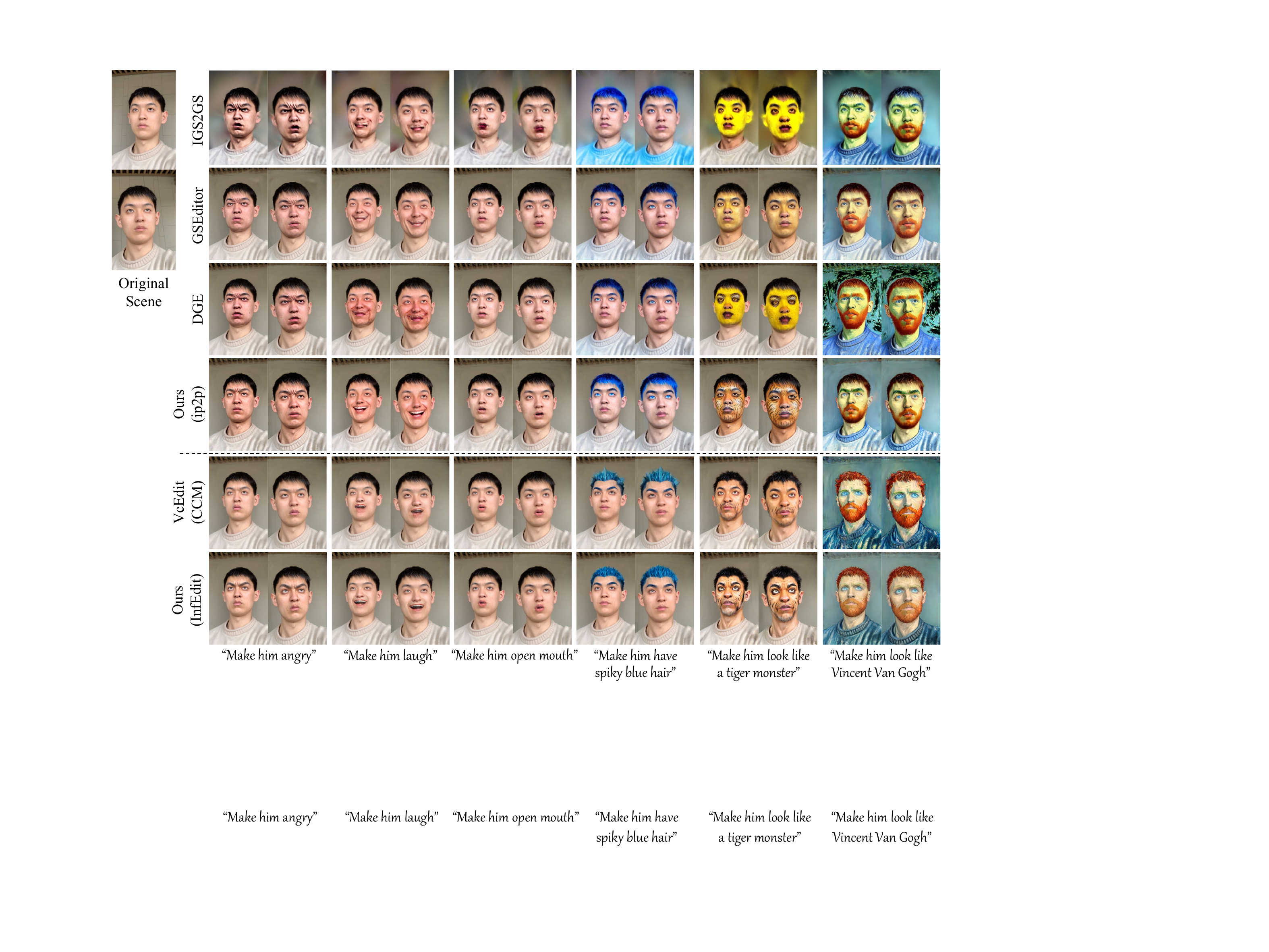} %
    \vspace{-.5cm}
    \caption{\textbf{Qualitative comparison} with IGS2GS~\cite{igs2gs}, GSEditor~\cite{chen2023gaussianeditorswiftcontrollable3d}, DGE~\cite{chen2024dgedirectgaussian3d} and VcEdit~\cite{wang2024viewconsistent3deditinggaussian} with CCM module. Our InterGSEdit framework achieves high-quality editing for both non-rigid and rigid tasks, demonstrating strong fidelity to textual instructions, and precise geometry preservation. 
    In contrast to other methods that produce edited results with tooth artifacts and unnatural modifications, our approach significantly reduces such artifacts, enhances texture quality, and produces more natural editing.
    }
    \label{fig3:experiment.}
    \vspace{-.3cm}
\end{figure*}
\section{Experiments}
\subsection{Experimental Settings}
\textbf{Implementation Details.} We implement our framework based on ThreeStudio~\cite{liu2023threestudio}. We utilize the diffusion model of InfEdit~\cite{xu2023inversionfreeimageeditingnatural} as the foundation for diffusion-based editing and employ the latent consistency model~\cite{luo2023latent} from HuggingFace. To demonstrate our method's capability in 3D model editing, we conduct experiments on various scenes from IN2N datasets~\cite{haque2023instructnerf2nerfediting3dscenes}. We represent 3D scenes using 3DGS~\cite{kerbl20233dgaussiansplattingrealtime} and leverage segmentation pipeline and semantic tracking of GSEditor~\cite{chen2023gaussianeditorswiftcontrollable3d} for local editing. Each editing is performed on a set of 20 random views, and the optimization of 3DGS is conducted for 800 to 1200 iterations depending on the complexity of the scene. All experiments are conducted on a NVIDIA RTX 4090 GPU.

\noindent \textbf{Evaluation Metrics.} We evaluate our method both qualitatively and quantitatively. The quantitative evaluation is based on the following metrics: CLIP Similarity~\cite{radford2021learningtransferablevisualmodels}, CLIP Text-Image Direction Similarity (CTIDS)~\cite{chenproedit}, and CLIP Direction Consistency (CDC)~\cite{haque2023instructnerf2nerfediting3dscenes}. 
Specifically, we randomly sample 20 camera views from the 3DGS training dataset for multi-view editing and fine-tune the 3DGS to obtain the final model with the edited images. For quantitative evaluation, we render multi-view images from the resulting 3D model using all available camera poses and assess the performance across the three evaluation metrics.
The CLIP similarity score is computed as the cosine similarity between the CLIP-encoded edited image embedding $E_{\mathrm{CLIP}}^{\mathrm{img}}(I_{\mathrm{edit}})$ and the target text embedding $E_{\mathrm{CLIP}}^{\mathrm{txt}}(T_{\mathrm{edit}})$. CTIDS~\cite{chenproedit} is calculated as the cosine similarity between the textual embedding difference $\Delta T=E_{\mathrm{CLIP}}^{\mathrm{txt}}(T_{\mathrm{edit}})-E_{\mathrm{CLIP}}^{\mathrm{txt}}(T_{\mathrm{src}})$ and the corresponding image embedding difference $\Delta I=E_{\mathrm{CLIP}}^{\mathrm{img}}(I_{\mathrm{edit}})-E_{\mathrm{CLIP}}^{\mathrm{img}}(I_{\mathrm{src}})$.
Additionally, following IN2N~\cite{haque2023instructnerf2nerfediting3dscenes}, CDC is used to measure directional consistency in editing results.

\subsection{Qualitative Comparisons}
\begin{figure*}[htp]
    \centering
    \includegraphics[width=1\textwidth]{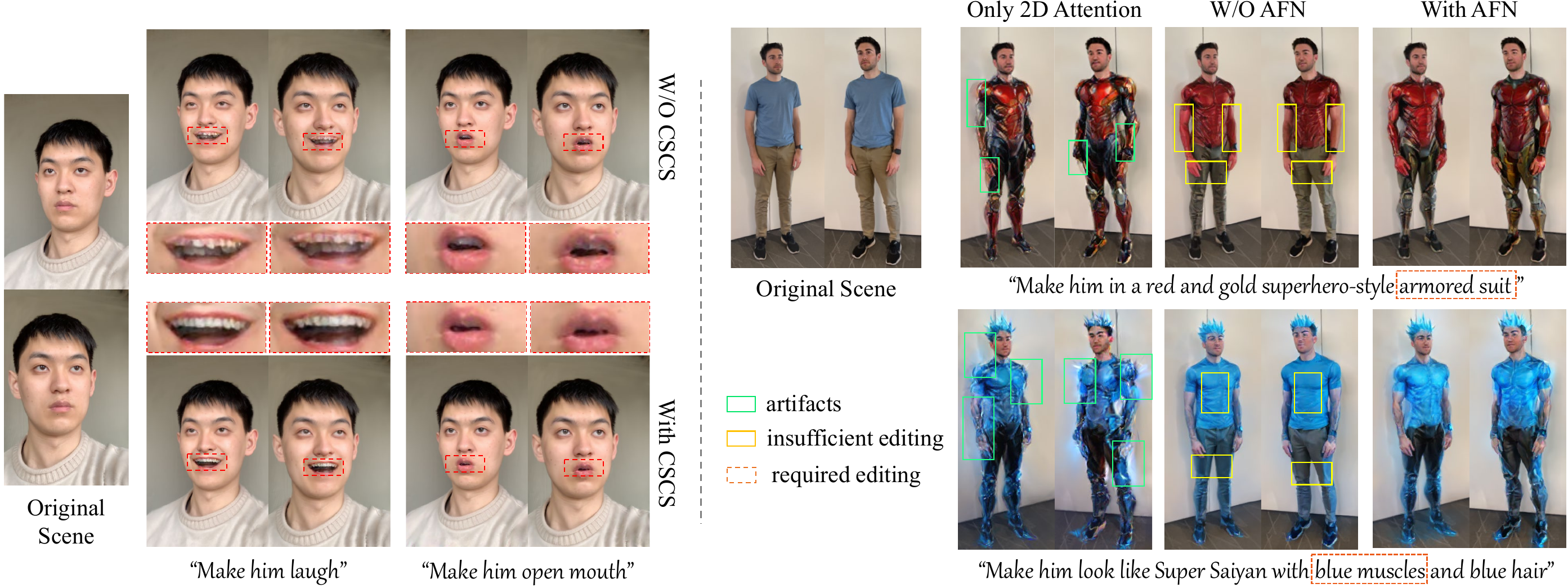} %
    \caption{\textbf{Ablation studies of our CLIP-based Semantic Consistency Selection (CSCS) and Attention Fusion Network (AFN).} 
     In the version ``W/O CSCS", non-rigid editing results, such as facial expression editing, exhibit serious tooth artifacts due to inconsistent features among different editing views. And the version ``W/O AFN" ignore the editing features of clothing, causing the humans are not edited properly and the texture details are not modified to match the textual description.
    }
    \label{fig4:ablation.}
\end{figure*}
To demonstrate the superiority of our InterGSEdit, we compare it with several baseline methods: IGS2GS~\cite{igs2gs}, GSEditor~\cite{chen2023gaussianeditorswiftcontrollable3d},  DGE~\cite{chen2024dgedirectgaussian3d} and VcEdit~\cite{wang2024viewconsistent3deditinggaussian} with CCM module. For IGS2GS, GSEditor, and DGE, we implement them with their released codes and models. Since the codes of VcEdit~\cite{wang2024viewconsistent3deditinggaussian} are not released publicly, we fulfill its cross-attention consistency module (CCM) and perform diffusion editing of InfEdit~\cite{xu2023inversionfreeimageeditingnatural} to compare with their CCM module on consistent editing.
Note that IGS2GS, GSEditor, and DGE adopt the InstructPix2Pix (ip2p)~\cite{brooks2023instructpix2pixlearningfollowimage} as their editing backbone, while VcEdit employs the InfEdit~\cite{xu2023inversionfreeimageeditingnatural}. To ensure a fair comparison and demonstrate the robustness of our approach across different diffusion frameworks, we conduct experiments using both ip2p and InfEdit for our method.
\cref{fig3:experiment.} presents qualitative comparisons in the Fangzhou scene. 

In the nonrigid cases, our InterGSEdit view-consistently produces more natural and realistic expressions, as demonstrated in the \textit{``make him angry"} and \textit{``make him laugh"} tasks. 
In contrast, existing methods tend to introduce undesirable artifacts, such as tooth artifacts, attributed to varying tooth features across multi-view editing results. 
For appearance and style transfer tasks, our method also outperforms other methods in editing quality. For example, when editing a man's hair to be blue and spiky hair, our method produces more natural colors and hair textures that accurately reflect the ``spiky" semantic characteristics.

\subsection{Quantitative Comparisons}
\begin{table}
  \centering
  \begin{tabular}{lccc}
    \toprule
    \textbf{Method} & \textbf{CLIP Similarity $\uparrow$} & \textbf{CTIDS $\uparrow$} & \textbf{CDC $\uparrow$} \\
    \midrule
    IGS2GS~\cite{igs2gs} & 0.2166 & 0.1328 & 0.8071 \\
    GSEditor~\cite{chen2023gaussianeditorswiftcontrollable3d} & 0.2169 & 0.1372 & 0.8099 \\
    DGE~\cite{chen2024dgedirectgaussian3d} & 0.2098 & 0.1097 & 0.8118 \\
    VcEdit~\cite{wang2024viewconsistent3deditinggaussian} & 0.2195 & 0.1195 & 0.8043 \\
    Ours (ip2p) & {0.2265} &{0.1416} & \textbf{0.8424} \\
    Ours (InfEdit)& \textbf{0.2285} & \textbf{0.1531} & {0.8347} \\
    \bottomrule
  \end{tabular}
  \caption{\textbf{Quantitative comparison} of different methods. Higher values indicate better performance. Notably, our InterGSEdit outperforms competing approaches in terms of CLIP similarity, CTIDS, and CDC metrics.}
  \label{tab:comparison}
  \vspace{-.3cm}
\end{table}

\begin{table}
  \centering
  \scalebox{0.92}{
    \begin{tabular}{lccc}
        \toprule
        \textbf{} & \textbf{CLIP Similarity $\uparrow$} & \textbf{CTIDS $\uparrow$} & \textbf{CDC $\uparrow$} \\
        \midrule
        W/O CSCS & 0.2034 & 0.1301 & 0.7647 \\
        With CSCS & \textbf{0.2090} & \textbf{0.1475} & \textbf{0.8218} \\
        \midrule
        Only 2D Attention & 0.2252 & \textbf{0.2835} & 0.8616 \\
        W/O AFN & 0.2206 & 0.1243 & 0.8429 \\
        With AFN & \textbf{0.2403} & 0.2738 & \textbf{0.8802} \\
        \bottomrule
    \end{tabular}
  }
    
  \caption{\textbf{Ablation study of our CLIP-based Semantic Consistency Selection (CSCS) and Attention Fusion Network (AFN).} The results demonstrate the effectiveness of CSCS and AFN in achieving high-quality and multi-view consistent editing results.}
  \label{tab:ablation}
  \vspace{-.3cm}
\end{table}
As shown in \cref{tab:comparison}, our method achieves the highest scores across all the metrics, demonstrating superior text-image alignment and multi-view consistency. Specifically, our method achieves a CLIP Similarity score of 0.2285, outperforming GSEditor (0.2169) and DGE (0.2098). Similarly, our method achieves a CTIDS score of 0.1531, surpassing GSEditor (0.1372) and DGE (0.1097), indicating that our approach better captures the semantic intent of the editing prompt. Additionally, our method attains the highest CDC score (0.8347), confirming improved 3D-aware editing consistency across different views.

\subsection{Ablation Study}
We evaluated the crucial role of CSCS and AFN within InterGSEdit. In the version without CSCS, a 3D attention prior is constructed by averaging unprojection from all views. 
As shown in \cref{fig4:ablation.}, ``W/O CSCS" results in obvious tooth artifacts due to geometry inconsistencies across different edited views. 
In the version without AFN, 3D-constrained attention is directly substituted for 2D cross-attention maps to generate the final outputs. 
And in the ``Only 2D Attention" version, 2D cross-attention features from the diffusion model remain unaltered during the editing process.
As shown in \cref{fig4:ablation.}, compared to the ``With AFN" version, the version ``W/O AFN" exhibits some regions  that are not fully edited due to the lack of original editing features, and the version ``Only 2D Attention" shows clear artifacts and blurred regions without the constraint of 3D geometry prior. Furthermore, without AFN, the much lower accuracy of the CTIDS metric indicates that the details are not recovered and the editing is not consistent to textual instructions. Notably, the ``Only 2D Attention" version shows a slightly higher CTIDS score than ours because this version utilizes full 2D cross-attention, leading to the most extensive editing. In contrast, our AFN fuses 3D constraint attention with 2D cross attention, making its CTIDS score theoretically a bit lower or equal.
Quantitatively, the improvements in CTIDS and CDC metrics indicate that our framework using CSCS and AFN not only aligns with the desired editing semantics, but also achieves multi-view consistency, as demonstrated in \cref{tab:ablation}.

\section{Conclusion}
In this work, we introduce InterGSEdit, an innovative 3D editing approach based on 3D Gaussian Splatting (3DGS) that enhances the quality and consisntency of 3D editing. In our framework, anchored with a user-specified key view, CLIP-based Semantic Consistency Selection is used to construct a 3D Geometry-Consistent Attention Piror from the selected reference views. In the subsequent multi-view editing process, 3D Geometry-Consistent Attention Piror is projected into 3D-constrained attention maps for all views, and then dynamically fused with the 2D cross-attention maps via the Attention Fusion Network. This fusing attention effectively guides the generation of editing diffusion to preserve geometric consistency and capture fine-grained details, producing coherent and realistic editing results. 

\section{Acknowledgments}
This work is supported in part by the National Natural Science Foundation of China under Grant 62272229 and 62472224, the Natural Science Foundation of Jiangsu Province under Grant BK20222012.

{
    \small
    \bibliographystyle{ieeenat_fullname}
    \bibliography{main}
}

\end{document}